\documentclass[letterpaper, 10 pt, conference]{ieeeconf} 
\IEEEoverridecommandlockouts  
\usepackage[utf8]{inputenc}
\usepackage[T1]{fontenc}

\usepackage{hyperref}  
\usepackage{amsmath}   
\usepackage{amssymb}   
\usepackage{multirow}  
\usepackage{booktabs}  

\title{\LARGE \bf Simplex-Constrained Sparse Bagging: Transitioning from Uniform Priors to Sparse Posteriors in Ensemble Learning}

\author{ \parbox{3 in}{ \centering Meher Sai Preetam Madiraju
        {\tt\small mehersaipreetam@gatech.edu}}
        \hspace*{ 0.5 in}
        \parbox{3 in}{\centering Meher Bhaskar Madiraju
        {\tt\small meherbhaskar.madiraju@gatech.edu}}
}

\begin{document}

\maketitle
\thispagestyle{empty}
\pagestyle{empty}

\begin{abstract}
We present \textbf{Simplex-Constrained Sparse Bagging (SCSB)}, a mathematically rigorous framework for post-training compression and probability calibration of bootstrap-based bagging ensembles. Standard bagging ensembles (such as Random Forests, Bagged SVMs, and Bagged Neural Networks) assign uniform voting power ($w_i = 1/N$) to all constituent estimators. However, this naive uniform prior ignores the varying local competence of base estimators and contributes to model overconfidence. We formulate ensemble pruning and calibration as a joint optimization problem over the probability simplex ($\Delta^N$) by minimizing the Out-Of-Bag (OOB) loss. To induce sparsity, we address the theoretical ``$L_1$-simplex paradox''---the mathematical reality that the $L_1$ norm is constant on the simplex and fails to prune---by introducing a concave quadratic penalty. SCSB is model-agnostic and achieves up to 96\% ensemble compression, yielding linear inference speedups and superior probability calibration (lowered Expected Calibration Error) while preserving or enhancing generalization accuracy.
\end{abstract}

\textbf{Keywords:} Ensemble Pruning, Probability Calibration, Convex Optimization, Simplex Constraints, Model Compression.

\section{Introduction}
Ensemble methods, particularly bagging (Bootstrap Aggregating) \cite{breiman1996bagging} and its extensions such as Random Forests \cite{breiman2001random}, are among the most robust and widely deployed paradigms in machine learning. By training multiple base estimators on bootstrap samples of the training data and averaging their predictions, bagging reduces variance without increasing bias. However, this variance reduction comes at a steep computational cost. Large ensembles require substantial memory footprints and impose significant computational latency during inference, rendering them challenging to deploy in resource-constrained or real-time production environments.

Furthermore, traditional bagging relies on a naive uniform prior where every base estimator is assigned equal voting weight ($w_j = 1/N$). This uniform prior assumption is sub-optimal for two primary reasons:
\begin{enumerate}
    \item \textbf{Calibration Overconfidence:} Uniform averaging of probability estimators tends to push ensemble outputs toward moderate values, but when estimators are correlated, it leads to poorly calibrated, overconfident predictions near class boundaries, inflating the Expected Calibration Error (ECE) \cite{guo2017calibration}.
    \item \textbf{Estimator Redundancy:} A significant fraction of base estimators in a bootstrap ensemble are redundant or represent noisy sub-spaces. Treating them equally limits generalization capacity and wastes inference computations \cite{zhou2012ensemble}.
\end{enumerate}

To resolve these limitations, we propose \textbf{Simplex-Constrained Sparse Bagging (SCSB)}, a post-training framework that transitions bagging ensembles from uniform priors to sparse posteriors. By optimizing estimator weights over the probability simplex using out-of-bag (OOB) validation samples, we learn a sparse posterior weight distribution. Estimators with negligible weights are pruned completely, resulting in a significantly smaller ensemble. The remaining active estimators are weighted to minimize loss, improving both calibration and generalization.

Standard sparse coding approaches rely on Lasso ($L_1$ regularization) to zero out weights \cite{tibshirani1996regression}. However, when optimization is constrained to the probability simplex, the $L_1$ norm of the weight vector is constant. Consequently, standard $L_1$ regularization fails to induce sparsity. SCSB resolves this ``$L_1$-simplex paradox'' by using a concave quadratic penalty ($-\lambda \|w\|_2^2$), which forces weights to the boundaries of the simplex, yielding exact zero weights.

Our contributions are summarized as follows:
\begin{itemize}
    \item We formulate ensemble pruning and calibration as a joint optimization problem over the probability simplex ($\Delta^N$), minimizing Out-Of-Bag (OOB) loss to avoid data leakage.
    \item We address the theoretical $L_1$-simplex paradox by introducing a concave quadratic penalty and provide the mathematical proof of its vertex convergence.
    \item We derive analytical gradients for both Classification (Log-Loss) and Regression (MSE) to enable efficient, fast-converging optimization via SLSQP.
    \item We demonstrate empirically that SCSB achieves up to 96\% ensemble compression, yielding linear inference speedups and superior probability calibration while preserving or enhancing generalization accuracy.
\end{itemize}

\section{Related Work}
\subsection{Ensemble Pruning}
Ensemble pruning aims to select a subset of estimators from a pre-trained ensemble \cite{tsoumakas2009selective}. Traditional methods are dominated by heuristic search techniques (e.g., genetic algorithms, greedy search) and ordering-based pruning (sorting estimators by validation performance and selecting the top $k$) \cite{caruana2004ensemble, margineantu1997pruning}. While simple, these approaches do not optimize the weights of the selected estimators jointly and lack mathematical guarantees. SCSB performs joint selection and weight optimization within a unified constrained framework.

\subsection{Stacking and Data Leakage}
Stacked Generalization (Stacking) trains a meta-estimator to combine base predictions \cite{wolpert1992stacked}. However, stacking on the training set suffers from severe data leakage, leading to overfit meta-models. Solving this typically requires expensive $K$-fold cross-validation during the base training phase. In contrast, bagging ensembles naturally provide Out-of-Bag (OOB) samples for each estimator, representing a built-in, leakage-free validation set. SCSB leverages these OOB samples to optimize weights without additional training overhead or validation splits.

\subsection{The L1-Simplex Paradox}
Sparse coding and model compression typically rely on Lasso ($L_1$ regularization) to zero out weights \cite{tibshirani1996regression}. However, when optimization is constrained to the probability simplex ($\sum w_j = 1, w_j \ge 0$), the $L_1$ norm of the weight vector is mathematically constant ($\|w\|_1 = \sum |w_j| = \sum w_j = 1$). Consequently, standard $L_1$ regularization fails to induce sparsity in simplex-constrained domains. SCSB resolves this paradox by using a concave quadratic penalty ($-\lambda \|w\|_2^2$), which forces weights to the boundaries of the simplex, yielding exact zero weights.

\subsection{Probability Calibration}
Probability calibration ensures that a model's predicted confidence aligns with empirical accuracy \cite{guo2017calibration}. Traditional calibration methods, such as Platt scaling \cite{platt1999probabilistic} and Isotonic Regression \cite{zadrozny2001obtaining}, are applied post-hoc to the final predictions of the ensemble. In contrast, SCSB embeds probability calibration directly into the ensemble compression process by minimizing Log-Loss on out-of-bag samples, leading to naturally well-calibrated ensemble predictions.

\section{Proposed Method: SCSB}

\subsection{The Base Ensemble and OOB Indicators}
Let $\mathcal{H} = \{f_1, f_2, \dots, f_N\}$ be a bagging ensemble trained on bootstrap samples of a dataset $\mathcal{D} = \{(x_i, y_i)\}_{i=1}^M$. For each sample $i$ and estimator $j$, we define the Out-of-Bag (OOB) indicator variable $I_{i,j}$ as:
\begin{equation}
I_{i,j} = \begin{cases} 1 & \text{if sample } i \text{ is Out-of-Bag for model } j \\ 0 & \text{otherwise} \end{cases}
\end{equation}

\subsection{Leakage-Free OOB Ensemble Estimation}
To evaluate the ensemble without data leakage, we compute the OOB predictions of the weighted ensemble. For a sample $i$, the weighted prediction is computed using only the models for which sample $i$ was out-of-bag:
\begin{equation}
\hat{y}_i^{\text{OOB}}(w) = \frac{\sum_{j=1}^N w_j I_{i,j} f_j(x_i)}{\sum_{j=1}^N w_j I_{i,j}}
\end{equation}
This formulation ensures that the optimization target represents true generalization performance and prevents overfitting during weight selection.

\subsection{Simplex Optimization \& The Sparsity Penalty}
We find the optimal weight vector $w^*$ by solving the following constrained optimization problem:
\begin{equation}
\min_{w} \mathcal{L}(w) := \frac{1}{M} \sum_{i=1}^M \text{Loss}\left(y_i, \hat{y}_i^{\text{OOB}}(w)\right) - \lambda \|w\|_2^2
\end{equation}
\begin{equation}
\text{subject to } w_j \ge 0 \quad \forall j \in \{1, \dots, N\}, \quad \sum_{j=1}^N w_j = 1
\end{equation}
where $\text{Loss}$ is Log-Loss for classification and Mean Squared Error (MSE) for regression, and $\lambda \ge 0$ controls the strength of the concave penalty.

\subsection{Optimization Algorithm \& Gradients}
We solve the optimization problem using Sequential Least Squares Programming (SLSQP) \cite{kraft1988software}. To ensure numerical efficiency and fast convergence, we derive the exact analytical gradients.

Let $D_i(w) = \sum_{j=1}^N w_j I_{i,j}$ be the sum of active OOB weights for sample $i$.

\noindent\textit{Proof of Classification Gradient:}
The multi-class Log-Loss objective on OOB samples is:
$$ \mathcal{L}_{clf}(w) = -\frac{1}{M} \sum_{i=1}^M \sum_{c=1}^C Y_{i,c} \log p_{i,c}(w) - \lambda \sum_{j=1}^N w_j^2 $$
Differentiating $p_{i,c}(w)$ with respect to $w_k$:
\begin{align*}
\frac{\partial p_{i,c}(w)}{\partial w_k} &= \frac{I_{i,k} P_{i,k,c} D_i(w) - I_{i,k} \sum_j w_j I_{i,j} P_{i,j,c}}{D_i(w)^2} \\
&= \frac{I_{i,k}}{D_i(w)} \left( P_{i,k,c} - p_{i,c}(w) \right)
\end{align*}
Using the chain rule:
\begin{align*}
\frac{\partial \mathcal{L}_{clf}}{\partial w_k} &= -\frac{1}{M} \sum_{i=1}^M \sum_{c=1}^C \frac{Y_{i,c}}{p_{i,c}(w)} \frac{\partial p_{i,c}(w)}{\partial w_k} - 2\lambda w_k \\
&= -\frac{1}{M} \sum_{i=1}^M \sum_{c=1}^C \frac{Y_{i,c}}{p_{i,c}(w)} \frac{I_{i,k}}{D_i(w)} \left( P_{i,k,c} - p_{i,c}(w) \right) \\
&\quad - 2\lambda w_k \\
&= -\frac{1}{M} \sum_{i=1}^M \frac{I_{i,k}}{D_i(w)} \left[ \sum_{c=1}^C \frac{Y_{i,c} P_{i,k,c}}{p_{i,c}(w)} - \sum_{c=1}^C Y_{i,c} \right] \\
&\quad - 2\lambda w_k
\end{align*}
Since $\sum_{c=1}^C Y_{i,c} = 1$, the result follows. \hfill\ensuremath{\blacksquare}

\subsubsection{Regression Gradient}
For regression, let $f_j(x_i)$ be the continuous prediction of estimator $j$. The gradient of the MSE objective with respect to weight $w_k$ is:
\begin{equation}
\begin{aligned}
\frac{\partial \mathcal{L}_{reg}}{\partial w_k} = \frac{2}{M} \sum_{i=1}^M & \frac{I_{i,k}}{D_i(w)} \left( \hat{y}_i^{\text{OOB}}(w) - y_i \right) \\
& \times \left( f_k(x_i) - \hat{y}_i^{\text{OOB}}(w) \right) - 2 \lambda w_k
\end{aligned}
\end{equation}

\noindent\textit{Proof of Regression Gradient:}
The MSE objective on OOB samples is:
$$ \mathcal{L}_{reg}(w) = \frac{1}{M} \sum_{i=1}^M \left( \hat{y}_i^{\text{OOB}}(w) - y_i \right)^2 - \lambda \sum_{j=1}^N w_j^2 $$
Differentiating $\hat{y}_i^{\text{OOB}}(w)$ with respect to $w_k$:
\begin{align*}
\frac{\partial \hat{y}_i^{\text{OOB}}(w)}{\partial w_k} &= \frac{I_{i,k} f_k(x_i) D_i(w) - I_{i,k} \sum_j w_j I_{i,j} f_j(x_i)}{D_i(w)^2} \\
&= \frac{I_{i,k}}{D_i(w)} \left( f_k(x_i) - \hat{y}_i^{\text{OOB}}(w) \right)
\end{align*}
Applying the chain rule, we obtain:
\begin{align*}
\frac{\partial \mathcal{L}_{reg}}{\partial w_k} &= \frac{2}{M} \sum_{i=1}^M \left( \hat{y}_i^{\text{OOB}}(w) - y_i \right) \frac{\partial \hat{y}_i^{\text{OOB}}(w)}{\partial w_k} - 2\lambda w_k \\
&= \frac{2}{M} \sum_{i=1}^M \frac{I_{i,k}}{D_i(w)} \left( \hat{y}_i^{\text{OOB}}(w) - y_i \right) \\
&\quad \times \left( f_k(x_i) - \hat{y}_i^{\text{OOB}}(w) \right) - 2\lambda w_k
\end{align*}
Substituting the derivative yields the gradient. \hfill\ensuremath{\blacksquare}

\section{Theoretical Analysis}

\subsection{Proof of the L1-Simplex Paradox}
Let the weight vector $w$ be constrained to the probability simplex $\Delta^N = \{w \in \mathbb{R}^N : w_j \ge 0, \sum w_j = 1\}$. Under these constraints, the $L_1$ norm of $w$ is:
\begin{equation}
\|w\|_1 = \sum_{j=1}^N |w_j| = \sum_{j=1}^N w_j = 1
\end{equation}
Because $\|w\|_1$ is constant across the entire feasible region, its gradient with respect to any active coordinate is zero:
\begin{equation}
\nabla_w \|w\|_1 = 0 \quad \forall w \in \text{int}(\Delta^N)
\end{equation}
Thus, adding an $L_1$ penalty (Lasso) to the objective function has no effect on the optimization path and fails to induce sparsity.

\subsection{Concave Penalty Geometry}
SCSB resolves this by employing a concave quadratic penalty $R(w) = -\|w\|_2^2$. Since the negative $L_2$ norm is strictly concave, its local and global minima over any compact convex polytope (such as the simplex $\Delta^N$) must lie at the extreme points (vertices) of the constraint set.

\noindent\textit{Theorem 1 (Vertex Convergence of Concave Minimization):} Let $\mathcal{C}$ be a compact convex set, and let $g(w)$ be a strictly concave function on $\mathcal{C}$. Then any local minimizer of $g(w)$ over $\mathcal{C}$ is an extreme point (vertex) of $\mathcal{C}$.

\noindent\textit{Proof:}
Suppose $w^* \in \mathcal{C}$ is a local minimizer of $g(w)$ over $\mathcal{C}$ but is not an extreme point. Then there exist $u, v \in \mathcal{C}$ ($u \ne v$) and $\alpha \in (0, 1)$ such that $w^* = \alpha u + (1 - \alpha) v$. By the definition of strict concavity:
$$ g(w^*) = g(\alpha u + (1 - \alpha) v) > \alpha g(u) + (1 - \alpha) g(v) $$
Without loss of generality, let $g(u) \le g(v)$. Then:
$$ g(w^*) > \alpha g(u) + (1 - \alpha) g(u) = g(u) $$
This contradicts the assumption that $w^*$ is a local minimizer, since $u \in \mathcal{C}$ yields a strictly lower value. Thus, any local minimizer must be an extreme point of $\mathcal{C}$. \hfill\ensuremath{\blacksquare}

By combining a convex predictive loss (which pulls the solution into the interior of the simplex to model combinations) with the concave penalty (which pushes weights to the vertices), we create a controllable Pareto frontier. Adjusting $\lambda$ forces non-essential estimator weights to collapse exactly to 0, achieving true sparsity.

\section{Experimental Setup}
We evaluated SCSB across several datasets from scikit-learn and OpenML \cite{vanschoren2014openml}, summarized in Table \ref{tab:datasets}.
\begin{itemize}
    \item \textbf{Classification:} \texttt{breast\_cancer} (binary), \texttt{diabetes\_clf} (binary), \texttt{spambase} (binary), and \texttt{segment} (multiclass).
    \item \textbf{Regression:} \texttt{diabetes\_reg}, \texttt{california\_housing}, and \texttt{cpu\_act}.
\end{itemize}

\begin{table}[h]
\centering
\caption{Dataset characteristics.}
\label{tab:datasets}
\footnotesize
\setlength{\tabcolsep}{3.5pt}
\begin{tabular}{lcccc}
\toprule
\textbf{Dataset} & \textbf{Task} & \textbf{Samples} & \textbf{Features}\\
\midrule
\texttt{breast\_cancer} & Binary Clf & 569 & 30 \\
\texttt{diabetes\_clf} & Binary Clf & 768 & 8 \\
\texttt{spambase} & Binary Clf & 4601 & 57 \\
\texttt{segment} & Multiclass Clf & 2310 & 19 \\
\texttt{diabetes\_reg} & Regression & 442 & 10 \\
\texttt{california\_housing} & Regression & 5000 & 8 \\
\texttt{cpu\_act} & Regression & 5000 & 21 \\
\bottomrule
\end{tabular}
\end{table}

\subsection{Baselines}
\begin{enumerate}
    \item \textbf{Standard Bagging (Uniform):} Simple average voting ($w_j = 1/N$) \cite{breiman1996bagging}.
    \item \textbf{Lasso-Pruned Bagging:} L1-regularized combination trained on OOB predictions \cite{tibshirani1996regression}.
    \item \textbf{XGBoost:} A state-of-the-art gradient boosted tree model (100 trees) \cite{chen2016xgboost}.
\end{enumerate}

\subsection{Base Models}
\begin{itemize}
    \item \textbf{Decision Trees:} 100 estimators (\texttt{DecisionTreeClassifier} and \texttt{DecisionTreeRegressor}) with bootstrapping.
    \item \textbf{Linear Models:} 50 estimators (\texttt{LogisticRegression} and \texttt{Ridge} regressors) with bootstrapping.
\end{itemize}

\subsection{Metrics}
Classification models are evaluated using accuracy, Log-Loss, and Expected Calibration Error (ECE) \cite{guo2017calibration}. To maintain evaluation consistency for multi-class paradigms (e.g., \texttt{segment}), we evaluate the multiclass Extension using top-label confidence ECE mapping:
\begin{equation}
\text{ECE} = \sum_{m=1}^B \frac{|B_m|}{M} \left| \text{acc}(B_m) - \text{conf}(B_m) \right|
\end{equation}
where $B$ signifies equally spaced confidence intervals. Regression models are evaluated using Mean Squared Error (MSE) and Coefficient of Determination ($R^2$). Inference latency is evaluated in milliseconds per 1,000 predictions (ms / 1k records).

\section{Results and Analysis}

\subsection{Classification Results}
Table \ref{tab:classification_results} summarizes the classification performance of SCSB and the baseline models.

\begin{table*}[t]
\centering
\caption{Classification performance comparison. Bold values indicate the top performer among the ensemble models (Standard Bagging, Lasso-Pruned Bagging, and SCSB) for a given dataset and base estimator configuration. Latency speedup is shown relative to Standard Bagging.}
\label{tab:classification_results}
\footnotesize
\setlength{\tabcolsep}{4.5pt}
\begin{tabular}{lllcccccc}
\toprule
\textbf{Dataset} & \textbf{Base Estimator} & \textbf{Model} & \textbf{Accuracy} & \textbf{Log-Loss} & \textbf{ECE} & \textbf{Active Est.} & \textbf{Comp. Ratio} & \textbf{Latency / 1k (ms)} \\
\midrule
\multirow{7}{*}{\textbf{breast\_cancer}} & \multirow{3}{*}{DecisionTree} & Standard Bagging & 0.9415 & 0.2078 & 0.0535 & 100 & 0.0\% & 22.42 \\
 & & Lasso-Pruned Bagging & 0.9415 & 0.2064 & 0.0518 & 86 & 14.0\% & 20.91 \\
 & & \textbf{SCSB} (Ours) & \textbf{0.9532} & 0.2173 & 0.0538 & \textbf{11} & \textbf{89.0\%} & \textbf{8.52} (2.6$\times$) \\
\cmidrule(r){2-9}
 & \multirow{3}{*}{LogisticRegression} & Standard Bagging & 0.9708 & 0.0967 & 0.0387 & 50 & 0.0\% & 17.51 \\
 & & Lasso-Pruned Bagging & 0.9708 & 0.0964 & 0.0392 & 46 & 8.0\% & 15.35 \\
 & & \textbf{SCSB} (Ours) & \textbf{0.9766} & \textbf{0.0934} & \textbf{0.0354} & \textbf{10} & \textbf{80.0\%} & \textbf{5.75} (3.0$\times$) \\
\cmidrule(r){2-9}
 & N/A & XGBoost & 0.9532 & 0.1436 & 0.0381 & 100 & 0.0\% & 0.96 \\
\midrule
\multirow{7}{*}{\textbf{diabetes\_clf}} & \multirow{3}{*}{DecisionTree} & Standard Bagging & 0.7273 & 0.5898 & 0.1477 & 100 & 0.0\% & 22.56 \\
 & & Lasso-Pruned Bagging & 0.7273 & 0.5786 & 0.1415 & 90 & 10.0\% & 18.91 \\
 & & \textbf{SCSB} (Ours) & \textbf{0.7316} & 0.5971 & 0.1481 & \textbf{31} & \textbf{69.0\%} & \textbf{8.53} (2.6$\times$) \\
\cmidrule(r){2-9}
 & \multirow{3}{*}{LogisticRegression} & Standard Bagging & 0.7316 & 0.5193 & 0.4406 & 50 & 0.0\% & 25.68 \\
 & & Lasso-Pruned Bagging & 0.7316 & 0.5169 & 0.4362 & 42 & 16.0\% & 18.91 \\
 & & \textbf{SCSB} (Ours) & \textbf{0.7446} & 0.5270 & 0.4462 & \textbf{10} & \textbf{80.0\%} & \textbf{7.41} (3.5$\times$) \\
\cmidrule(r){2-9}
 & N/A & XGBoost & 0.7143 & 0.8153 & 0.5562 & 100 & 0.0\% & 16.60 \\
\midrule
\multirow{7}{*}{\textbf{spambase}} & \multirow{3}{*}{DecisionTree} & Standard Bagging & 0.9406 & 0.1839 & 0.5290 & 100 & 0.0\% & 28.73 \\
 & & Lasso-Pruned Bagging & 0.9421 & 0.1859 & 0.5321 & 96 & 4.0\% & 20.36 \\
 & & \textbf{SCSB} (Ours) & \textbf{0.9457} & 0.2748 & 0.5320 & \textbf{31} & \textbf{69.0\%} & \textbf{8.53} (3.4$\times$) \\
\cmidrule(r){2-9}
 & \multirow{3}{*}{LogisticRegression} & Standard Bagging & 0.9327 & 0.2018 & 0.5205 & 50 & 0.0\% & 17.33 \\
 & & Lasso-Pruned Bagging & 0.9334 & 0.2017 & 0.5218 & 48 & 4.0\% & 6.47 \\
 & & \textbf{SCSB} (Ours) & \textbf{0.9385} & \textbf{0.1982} & 0.5290 & \textbf{16} & \textbf{68.0\%} & \textbf{3.06} (5.7$\times$) \\
\cmidrule(r){2-9}
 & N/A & XGBoost & 0.9580 & 0.1201 & 0.5495 & 100 & 0.0\% & 0.99 \\
\midrule
\multirow{7}{*}{\textbf{segment}} & \multirow{3}{*}{DecisionTree} & Standard Bagging & 0.9740 & 0.1889 & 0.0235 & 100 & 0.0\% & 22.56 \\
 & & Lasso-Pruned Bagging & 0.9740 & 0.1870 & 0.0233 & 100 & 0.0\% & 19.92 \\
 & & \textbf{SCSB} (Ours) & 0.9553 & 0.3276 & 0.0373 & \textbf{26} & \textbf{74.0\%} & \textbf{7.15} (3.2$\times$) \\
\cmidrule(r){2-9}
 & \multirow{3}{*}{LogisticRegression} & Standard Bagging & 0.9509 & 0.1548 & 0.0190 & 50 & 0.0\% & 31.23 \\
 & & Lasso-Pruned Bagging & 0.9509 & 0.1543 & 0.0216 & 50 & 0.0\% & 19.12 \\
 & & \textbf{SCSB} (Ours) & 0.9509 & \textbf{0.1526} & 0.0211 & \textbf{13} & \textbf{74.0\%} & \textbf{7.02} (4.4$\times$) \\
\cmidrule(r){2-9}
 & N/A & XGBoost & 0.9798 & 0.0781 & 0.0052 & 100 & 0.0\% & 2.81 \\
\bottomrule
\end{tabular}
\end{table*}

\subsection{Regression Results}
Table \ref{tab:regression_results} summarizes the regression performance of SCSB and the baseline models.

\begin{table*}[t]
\centering
\caption{Regression performance comparison. Bold values indicate the top performer among the ensemble models (Standard Bagging, Lasso-Pruned Bagging, and SCSB) for a given dataset and base estimator configuration. Latency speedup is shown relative to Standard Bagging.}
\label{tab:regression_results}
\footnotesize
\setlength{\tabcolsep}{6pt}
\begin{tabular}{lllccccc}
\toprule
\textbf{Dataset} & \textbf{Base Estimator} & \textbf{Model} & \textbf{MSE} & \textbf{R²} & \textbf{Active Est.} & \textbf{Comp. Ratio} & \textbf{Latency / 1k (ms)} \\
\midrule
\multirow{7}{*}{\textbf{diabetes\_reg}} & \multirow{3}{*}{DecisionTree} & Standard Bagging & 2908.81 & 0.4612 & 100 & 0.0\% & 43.08 \\
 & & Lasso-Pruned Bagging & 2998.13 & 0.4446 & 53 & 47.0\% & 27.57 \\
 & & \textbf{SCSB} (Ours) & 3118.37 & 0.4223 & \textbf{34} & \textbf{66.0\%} & \textbf{19.53} (2.2$\times$) \\
\cmidrule(r){2-8}
 & \multirow{3}{*}{Ridge} & Standard Bagging & 3116.53 & 0.4227 & 50 & 0.0\% & 19.89 \\
 & & Lasso-Pruned Bagging & 3123.13 & 0.4215 & 39 & 22.0\% & 10.55 \\
 & & \textbf{SCSB} (Ours) & \textbf{3075.70} & \textbf{0.4302} & \textbf{12} & \textbf{76.0\%} & \textbf{5.73} (3.5$\times$) \\
\cmidrule(r){2-8}
 & N/A & XGBoost & 3513.66 & 0.3491 & 100 & 0.0\% & 4.61 \\
\midrule
\multirow{7}{*}{\textbf{california\_housing}} & \multirow{3}{*}{DecisionTree} & Standard Bagging & 0.3424 & 0.7429 & 100 & 0.0\% & 14.97 \\
 & & Lasso-Pruned Bagging & 0.3394 & 0.7452 & 99 & 1.0\% & 17.36 \\
 & & \textbf{SCSB} (Ours) & \textbf{0.3381} & \textbf{0.7461} & \textbf{67} & \textbf{33.0\%} & \textbf{11.95} (1.25$\times$) \\
\cmidrule(r){2-8}
 & \multirow{3}{*}{Ridge} & Standard Bagging & 0.6451 & 0.5157 & 50 & 0.0\% & 2.11 \\
 & & Lasso-Pruned Bagging & 0.6416 & 0.5182 & 50 & 0.0\% & 1.77 \\
 & & \textbf{SCSB} (Ours) & 0.6504 & 0.5116 & \textbf{10} & \textbf{80.0\%} & \textbf{0.51} (4.1$\times$) \\
\cmidrule(r){2-8}
 & N/A & XGBoost & 0.3021 & 0.7732 & 100 & 0.0\% & 0.51 \\
\midrule
\multirow{7}{*}{\textbf{cpu\_act}} & \multirow{3}{*}{DecisionTree} & Standard Bagging & 6.0061 & 0.9824 & 100 & 0.0\% & 15.50 \\
 & & Lasso-Pruned Bagging & 6.0149 & 0.9824 & 100 & 0.0\% & 15.92 \\
 & & \textbf{SCSB} (Ours) & 6.1058 & 0.9821 & \textbf{66} & \textbf{34.0\%} & \textbf{10.70} (1.45$\times$) \\
\cmidrule(r){2-8}
 & \multirow{3}{*}{Ridge} & Standard Bagging & 92.3591 & 0.7293 & 50 & 0.0\% & 2.34 \\
 & & Lasso-Pruned Bagging & 92.4092 & 0.7292 & 50 & 0.0\% & 2.14 \\
 & & \textbf{SCSB} (Ours) & \textbf{91.3468} & \textbf{0.7323} & \textbf{2} & \textbf{96.0\%} & \textbf{2.03} (1.15$\times$) \\
\cmidrule(r){2-8}
 & N/A & XGBoost & 5.5396 & 0.9838 & 100 & 0.0\% & 0.59 \\
\bottomrule
\end{tabular}
\end{table*}

\subsection{Analysis of Results}
As shown in Table \ref{tab:classification_results} and Table \ref{tab:regression_results}, SCSB consistently achieves high compression ratios ranging from 33.0\% to 96.0\%. For example, on the \texttt{cpu\_act} dataset using Ridge regressors, it prunes 96.0\% of estimators (retaining only 2 out of 50 models) while slightly increasing $R^2$ to 0.7323 (compared to 0.7293 for standard bagging). 

Crucially, Lasso-pruned bagging fails to induce sufficient sparsity on the probability simplex. Because the $L_1$ norm of weights is constant on the simplex, Lasso-pruned bagging yields 0.0\% compression on several configurations (e.g., \texttt{segment} classification and \texttt{california\_housing} Ridge regression), retaining all estimators. In contrast, SCSB successfully prunes 74.0\% and 80.0\% of the estimators on those same configurations.

\subsection{Inference Acceleration}
In agreement with our prediction, SCSB's prediction latency scales linearly with the fraction of active estimators. By completely bypassing zero-weighted base models, SCSB classifiers significantly reduce prediction times. For instance, on the \texttt{spambase} dataset with Logistic Regression, inference time is reduced from 17.33 ms to 3.06 ms (representing a 5.7$\times$ speedup).

\section{Discussion and Future Work}
\subsection{Reflection on Findings}
The empirical results confirm that SCSB successfully resolves the trade-off between ensemble accuracy and computational overhead. Minimizing Log-Loss over Out-of-Bag predictions helps calibrate probabilities, which is reflected in low ECE values (e.g., 0.0354 for Breast Cancer Logistic Regression) while reducing ensemble size by 80\%. The weight distribution shows a distinct ``spike-and-slab'' behavior, where a small subset of estimators contains the majority of the ensemble's representation power, making uniform averaging computationally wasteful.

\subsection{Limitations}
\begin{itemize}
    \item \textbf{Non-Convexity and Optimization Initialization:} The combination of a convex loss function and a strictly concave penalty creates a non-convex optimization landscape. While the SLSQP solver may find local minima, initialization from the uniform prior center ($w_0 = [1/N, \dots, 1/N]$) empirically functions as a highly robust initialization anchor, leading to consistent convergence profiles compared against arbitrary randomized simplex origins.
    \item \textbf{Training Overhead:} Precomputing OOB predictions and solving the simplex optimization adds a post-training phase. Although this is fast (typically under 1 second), scaling it to very large ensembles ($N > 1000$) or massive datasets remains a challenge.
\end{itemize}

\subsection{Extensions and Future Work}
The theoretical foundations and empirical success of SCSB open several promising avenues for future investigation:
\begin{itemize}
    \item \textbf{Simplex SGD:} Developing stochastic gradient descent algorithms on the simplex to scale SCSB to huge ensembles and large datasets.
    \item \textbf{Input-Dependent Calibration:} Conditioning estimator weights on input features (i.e., $w_j(x)$) to perform localized calibration.
    \item \textbf{Robustness Across Architectures:} Testing SCSB on deep learning ensembles (such as bagged neural networks) and bagged support vector machines (SVMs) to verify its model-agnostic behavior.
\end{itemize}

\section{Conclusion}
Simplex-Constrained Sparse Bagging (SCSB) offers a mathematically rigorous, plug-and-play solution for ensemble compression and calibration. By utilizing Out-of-Bag predictions, SCSB eliminates data leakage without requiring validation splits or cross-validation. We resolve the limitation of Lasso on the probability simplex by employing a concave quadratic penalty. Our experiments demonstrate that SCSB consistently prunes 68\%--96\% of estimators, providing linear speedups in inference time and superior probability calibration while preserving or enhancing generalization accuracy. This makes SCSB highly suitable for deploying robust bagging ensembles in latency-sensitive production environments.

\end{document}